\documentclass[conference,flushend]{iaria} 

\usepackage{amsmath} 

\usepackage{graphicx} 

\usepackage{booktabs}
\usepackage{tabularx}
\usepackage{subfig}

\title{Physics-Informed Neural Network Surrogate Models for River Stage Prediction}

\author{
  \IEEEauthorblockN{Maximilian Zoch}
  \IEEEauthorblockA{\textit{CoDiS-Lab ISDS} \\
    \textit{Graz Technical University of Technology} \\
    Graz, Austria \\
    maximilian.zoch@tugraz.at}
  \and
  \IEEEauthorblockN{Edward Holmberg}
  \IEEEauthorblockA{\textit{Gulf States Center for Environmental Informatics} \\
    \textit{University of New Orleans} \\
    Louisiana, United States \\
    eholmber@uno.edu}
  \and
  \IEEEauthorblockN{Pujan Pokhrel}
  \IEEEauthorblockA{\textit{Gulf States Center for Environmental Informatics} \\
    \textit{University of New Orleans} \\
    Louisiana, United States \\
    ppokhre1@uno.edu}
  \and
  \IEEEauthorblockN{Ken Pathak}
  \IEEEauthorblockA{\textit{US Army Corps of Engineers} \\
    \textit{Vicksburg District} \\
    Mississippi, United States \\
    ken.pathak@usace.army.mil}
  \and
  \IEEEauthorblockN{Steven Sloan}
  \IEEEauthorblockA{\textit{US Army Corps of Engineers} \\
    \textit{Vicksburg District} \\
    Mississippi, United States \\
    steven.sloan@usace.army.mil}
  \and
  \IEEEauthorblockN{Kendall Niles}
  \IEEEauthorblockA{\textit{US Army Corps of Engineers} \\
    \textit{Vicksburg District} \\
    Mississippi, United States \\
    kendall.niles@usace.army.mil}
  \and
  \IEEEauthorblockN{Jay Ratcliff}
  \IEEEauthorblockA{\textit{Gulf States Center for Environmental Informatics} \\
    \textit{University of New Orleans} \\
    Louisiana, United States \\
    jratclif@uno.edu}
  \and
  \IEEEauthorblockN{Maik Flanagin}
  \IEEEauthorblockA{\textit{US Army Corps of Engineers} \\
    \textit{New Orleans District} \\
    Louisiana, United States \\
    maik.c.flanagin@usace.army.mil}
  \and
  \IEEEauthorblockN{Elias Ioup}
  \IEEEauthorblockA{\textit{Center for Geospatial Sciences} \\
    \textit{Naval Research Laboratory} \\
    Mississippi, United States \\
    elias.z.ioup.civ@us.navy.mil}
  \and
  \IEEEauthorblockN{Christian Guetl}
  \IEEEauthorblockA{\textit{CoDiS-Lab ISDS} \\
    \textit{Graz Technical University of Technology} \\
    Graz, Austria \\
    c.guetl@tugraz.at}
  \and
  \IEEEauthorblockN{Mahdi Abdelguerfi}
  \IEEEauthorblockA{\textit{Gulf States Center for Environmental Informatics} \\
    \textit{University of New Orleans} \\
    Louisiana, United States \\
    gulfsceidirector@uno.edu}
}

\begin{document}

\maketitle
\begin{abstract}
This work investigates the feasibility of using Physics-Informed Neural Networks (PINNs) as surrogate models for river stage prediction, aiming to reduce computational cost while maintaining predictive accuracy. Our primary contribution demonstrates that PINNs can successfully approximate HEC-RAS numerical solutions when trained on a single river, achieving strong predictive accuracy with generally low relative errors, though some river segments exhibit higher deviations.

By integrating the governing Saint-Venant equations into the learning process, the proposed PINN-based surrogate model enforces physical consistency and significantly improves computational efficiency compared to HEC-RAS. We evaluate the model’s performance in terms of accuracy and computational speed, demonstrating that it closely approximates HEC-RAS predictions while enabling real-time inference.

These results highlight the potential of PINNs as effective surrogate models for single-river hydrodynamics, offering a promising alternative for computationally efficient river stage forecasting. Future work will explore techniques to enhance PINN training stability and robustness across a more generalized multi-river model.
\end{abstract}


\section{Introduction}

Rivers and waterways play a critical role in sustaining agricultural, industrial, and urban infrastructure. Understanding river stage dynamics is essential for a wide range of applications, including crop irrigation, drinking water supply, drainage planning, and flood risk assessment \cite{chanson2004, usace-manual}. Accurate river stage prediction enables informed decision-making in these domains, with economic, environmental, and societal benefits. During extreme weather events such as hurricanes or heavy rainfall, the ability to make real-time predictions of river behavior is particularly crucial for flood forecasting and emergency response \cite{ipcc2022, henley2024}.

Traditional hydrodynamic models such as the Hydrologic Engineering Center’s River Analysis System (HEC-RAS) provide high-fidelity simulations of river stage by solving the Saint-Venant equations \cite{usace-manual, chanson2004}. While these models are widely used for flood risk analysis, they are computationally expensive, requiring extensive parameter calibration and fine spatial and temporal resolution. As a result, simulating future water levels can take hours or days, making real-time forecasting infeasible in rapidly evolving flood scenarios.\cite{malcherek2019}.

To address this challenge, this work investigates the development of a \textit{physics-informed surrogate model} for river stage prediction. Unlike purely data-driven approaches, \textit{physics-informed neural networks (PINNs)} integrate the governing \textit{Saint-Venant equations} into the learning process, enforcing physics-based constraints while improving model generalization beyond the training domain \cite{raissi2019, feng2023}. This approach enables more \textit{computationally efficient} predictions while maintaining \textit{physical consistency}.

This study focuses on developing and evaluating a \textit{single-river PINN-based surrogate model} that approximates HEC-RAS river stage predictions. The primary objectives are:

\begin{itemize}
    \item \textbf{Assessing the accuracy and computational efficiency} of a physics-informed surrogate model trained on a single river.
    \item \textbf{Comparing PINN predictions to HEC-RAS outputs} to determine the feasibility of using surrogate modeling for river hydrodynamics.
    \item \textbf{Identifying challenges and limitations} in training PINNs for river stage prediction, establishing a foundation for future work in extending these models to multiple river systems.
\end{itemize}

By demonstrating the feasibility of physics-informed surrogate models for single-river applications, this work aims to provide a stepping stone for future research into broader hydrodynamic modeling frameworks.

\section{Background and Related Work}

\subsection{HEC-RAS: A Computational Numerical River Model}
The \textit{Hydrologic Engineering Center’s River Analysis System (HEC-RAS)}, developed by the \textit{U.S. Army Corps of Engineers (USACE)}, is an \textit{industry-standard numerical model} for simulating open-channel flow \cite{usace-manual}. HEC-RAS is widely used in flood forecasting, infrastructure planning, and water resource management \cite{usace-hydraulic}.

At its core, HEC-RAS numerically solves the \textit{Saint-Venant equations}, a system of \textit{shallow water PDEs} that govern mass and momentum conservation in river channels \cite{chanson2004}. The \textit{1D Saint-Venant equations} are given by:

\begin{align}
    \frac{\partial A}{\partial t} + \frac{\partial (Au)}{\partial x} &= 0,  \quad \text{(Continuity Equation)}
    \label{eq:continuity}
\end{align}
\begin{align}
    \frac{\partial u}{\partial t} + u \frac{\partial u}{\partial x} + g \frac{\partial h}{\partial x} + S_f - S_0 &= 0,  \quad \text{(Momentum Equation)}
    \label{eq:momentum}
\end{align}

where:
\begin{itemize}
    \item \( A(x,t) \) is the cross-sectional flow area,
    \item \( u(x,t) \) is velocity,
    \item \( h(x,t) \) is water surface elevation,
    \item \( S_f \) is the friction slope,
    \item \( S_0 \) is the bed slope.
\end{itemize}

\subsubsection{HEC-RAS Inputs}

\begin{itemize}
    \item \textbf{Geometric Data}: Cross-sectional profiles of riverbanks and channel bottoms, which are often stored in geospatial databases and used for hydrodynamic simulations \cite{MFlanagin2007, Chung2001}.

    \item \textbf{Boundary Conditions}: Flow rates, water levels, and upstream/downstream conditions.
    \item \textbf{Hydraulic Parameters}: Manning’s roughness coefficients, channel slopes, and obstructions \cite{flanagin2007}.
\end{itemize}

\subsubsection{HEC-RAS Outputs}

\begin{itemize}
    \item \textbf{Stage Predictions}: Water surface elevations over time.
    \item \textbf{Flow Predictions}: Velocity distribution across river stations.
\end{itemize}

While HEC-RAS provides high-fidelity results, it is \textit{computationally expensive}, requiring iterative solvers and extensive parameter calibration \cite{malcherek2019}. This computational burden makes real-time forecasting impractical in flood response scenarios.

\subsection{Surrogate Models in Computational Science}
To mitigate computational costs, \textit{surrogate models} approximate numerical solvers by learning the relationship between \textit{input parameters} (e.g., river geometry, boundary conditions) and \textit{output predictions} (e.g., water surface elevation) without directly solving PDEs. These models are trained on diverse input-output pairs from numerical simulations, enabling them to predict approximate solutions at significantly reduced computational cost \cite{benner2015}.

Surrogate models have been successfully applied in \textit{fluid dynamics, aerodynamics, and weather prediction}, demonstrating their ability to reduce the complexity of PDE-based simulations \cite{bi2023}. However, purely data-driven surrogate models, such as artificial neural networks (ANNs) and Gaussian processes, \textit{lack physical consistency}, leading to poor generalization when applied to dynamic, unseen river conditions \cite{takbiri2020}.

\subsection{Physics-Informed Neural Networks (PINNs)}
Physics-Informed Neural Networks (PINNs) provide an alternative approach by \textit{embedding governing physics equations into their training process}. Unlike traditional surrogate models that rely solely on input-output mappings, PINNs enforce \textit{physical laws} (e.g., conservation of mass and momentum) as constraints in their loss function \cite{raissi2019}. 

By minimizing residuals from PDEs during training, PINNs generate solutions that remain \textit{consistent with known physics}, even in \textit{data-scarce environments}. PINNs have demonstrated effectiveness in \textit{computational fluid dynamics (CFD)}, hydrodynamic simulators, and geophysics \cite{donnelly2024, bai2020}, but \textit{their application in river stage prediction remains limited} \cite{feng2023}. 

\subsection{Fourier Feature Encoding in Surrogate Models}
Machine learning applications in fluid dynamics often employ \textit{Fourier feature encoding} to improve the model’s ability to learn fine-scale variations in spatial and temporal data \cite{mildenhall2020nerf}. Standard neural networks exhibit a spectral bias toward learning low-frequency functions \cite{rahaman2019spectral}, which can lead to poor generalization in high-variability physical systems.

Fourier feature encoding mitigates this bias by transforming input coordinates into a high-dimensional space:

\begin{equation}
\gamma(x) = \left[ \cos(2\pi B x), \sin(2\pi B x) \right]^T
\end{equation}

where \( B \) is a matrix of random Fourier base frequencies. This transformation enables neural networks to capture fine-grained variations in river stage predictions.

\subsection{Research Gap and Motivation for Our Approach}
Despite advancements in surrogate modeling and physics-informed learning, \textit{PINNs have not been widely applied as surrogate models for river stage prediction}. Previous work on PINNs has primarily focused on \textit{idealized fluid simulations} rather than real-world hydrodynamic systems governed by HEC-RAS data.

This work addresses these gaps by:
\begin{itemize}
    \item Developing a \textit{PINN-based surrogate model} for river stage prediction that enforces the \textit{Saint-Venant equations} as physical constraints.
    \item Investigating the effectiveness of Fourier feature encoding in improving the model’s ability to capture fine-scale variations in river flow.
    \item Evaluating whether \textit{our physics-informed surrogate model} can achieve accuracy comparable to HEC-RAS while significantly reducing computational cost.
\end{itemize}

The following sections describe our proposed methodology in detail.

\section{Methodology}

\subsection{Problem Formulation}
The primary objective of this study is to develop a computationally efficient surrogate model for river \textit{stage prediction}, denoted as \(h(x,t)\), using a PINN. The model is trained on simulated river data from the HEC-RAS, a numerical solver developed by the USACE. HEC-RAS provides high-fidelity water surface elevation predictions by solving the \textit{Saint-Venant equations}, but its computational complexity makes real-time forecasting infeasible. Our approach seeks to approximate the HEC-RAS stage predictions while significantly reducing inference time.

Given a river cross-section and boundary conditions, the proposed surrogate model minimizes:

\begin{equation}
\mathcal{F}(h, u, A) = 0,
\end{equation}

where:
\begin{itemize}
    \item \( h(x,t) \) is the \textit{water surface elevation},
    \item \( u(x,t) \) is the \textit{flow velocity},
    \item \( A(x,t) \) is the \textit{cross-sectional area}.
\end{itemize}

The function \( \mathcal{F} \) represents the \textbf{Saint-Venant equations}, ensuring that predictions adhere to known physical constraints.

\begin{figure}[htbp]
    \centering
    \includegraphics[width=0.85\linewidth]{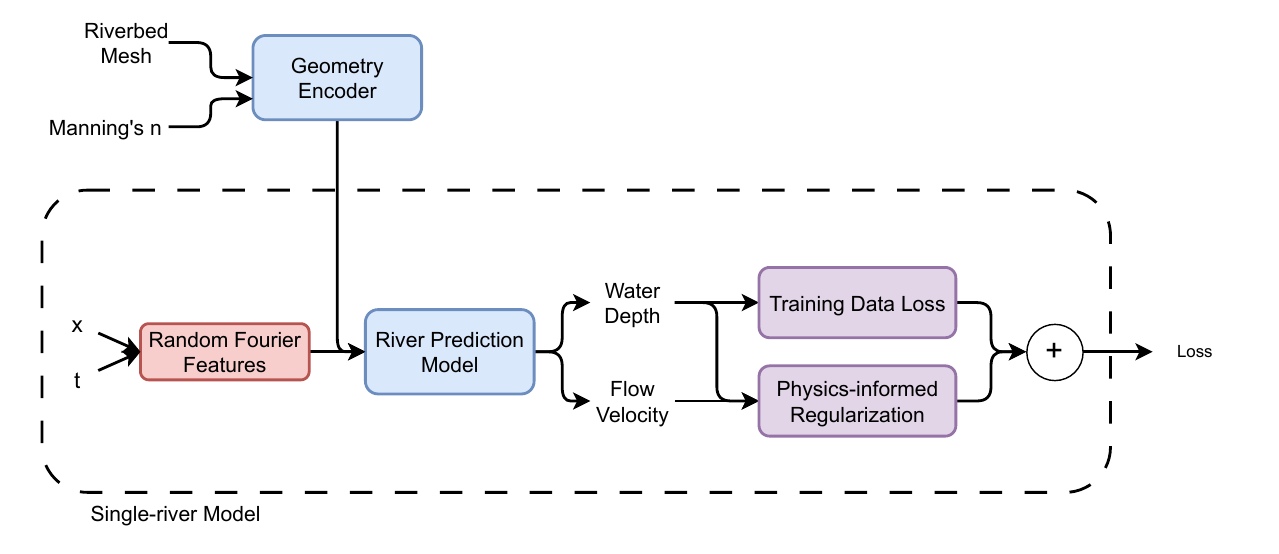}
    \caption{System overview of the single-river PINN surrogate model.}
    \label{fig:system-overview}
\end{figure}

\subsection{Surrogate Model Architecture}
The surrogate model consists of a deep neural network that approximates HEC-RAS river stage predictions while incorporating physics-informed regularization. The architecture follows a supervised learning approach with additional physics-based constraints to ensure compliance with governing hydrodynamic equations.

\subsubsection{Fourier Feature Encoding}
Neural networks typically exhibit a bias toward learning low-frequency functions \cite{rahaman2019spectral}. To mitigate this and improve fine-scale resolution, \textbf{Fourier feature encoding} is applied to the input coordinates:

\begin{equation}
\gamma(x) = \left[ \cos(2\pi B x), \sin(2\pi B x) \right]^T
\end{equation}

where \( B \) is a matrix of random Fourier base frequencies \cite{mildenhall2020nerf}. This transformation enhances the model’s ability to capture complex variations in river stage across time and space. The standard deviation \( \sigma \) of \( B \) is optimized through grid search to balance spectral bias and overfitting.

\subsubsection{Neural Network Architecture}
The surrogate model adopts an implicit neural representation, mapping spatial and temporal inputs \((x,t)\) to predicted river stage \(h(x,t)\) and flow velocity \(u(x,t)\). The architecture is structured as follows:

\begin{itemize}
    \item \textbf{Input Layer:} Encodes river mile \( x \) and time \( t \) using Fourier features.
    \item \textbf{Hidden Layers:} 6 fully connected residual blocks with 512 hidden dimensions per layer.
    \item \textbf{Output Layer:} Predicts \( h(x,t) \) (water depth) and \( u(x,t) \) (flow velocity).
\end{itemize}

\begin{figure}[htbp]
    \centering
    \includegraphics[width=0.5\linewidth]{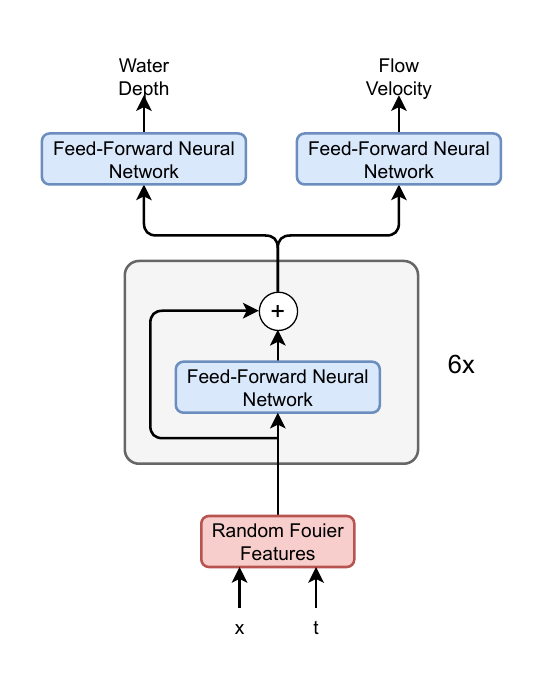}
    \caption{Structure of the single-river surrogate model.}
    \label{fig:single-river-model}
\end{figure}

Residual connections are used to improve training stability and convergence \cite{He2015ResNet}. Figure \ref{fig:single-river-model} depicts the architecture of the single-river model.

\subsection{Loss Function}
The surrogate model is trained using a hybrid loss function combining supervised learning and physics-informed regularization:

\begin{equation}
\mathcal{L} = \mathcal{L}_{\text{HEC-RAS}} + \lambda \mathcal{L}_{\text{physics}},
\end{equation}

where:
\begin{itemize}
    \item \( \mathcal{L}_{\text{HEC-RAS}} \) is the \textit{supervised loss}, measuring deviation from \textit{HEC-RAS-generated stage predictions}.
    \item \( \mathcal{L}_{\text{physics}} \) enforces compliance with the \textit{Saint-Venant equations}, ensuring physically valid water surface elevations.
\end{itemize}

The physics-informed term is derived from the continuity and momentum equations:

\begin{equation}
    \mathcal{L}_{\text{physics}} = \left\| \frac{\partial h}{\partial t} + \frac{\partial (h u)}{\partial x} \right\|^2 + \left\| \frac{\partial u}{\partial t} + u \frac{\partial u}{\partial x} + g \frac{\partial h}{\partial x} \right\|^2.
\end{equation}

These residual terms are computed using automatic differentiation in PyTorch.

\section{Experimental Setup}

\subsection{Dataset Description}
The dataset used in this study represents various river segments of the Mississippi River. In our research, each segment is treated as a single-river model.
It consists of:
\begin{itemize}
    \item 63 river segments spanning a significant portion of the Mississippi River,
    \item 3,240 river stations strategically placed for high-fidelity hydrodynamic simulation.
\end{itemize}

These stations were positioned by river modelers to capture key variations in water surface elevation and flow characteristics while minimizing redundant cross-sections. On average, stations are spaced \textbf{0.74 miles apart}, with time-series data generated for \textbf{1D unsteady flow analysis}. 

\begin{table}[htbp]
    \centering
    \caption{Dataset Overview for Benchmarking Study}
    \small
    \begin{tabular}{l l} 
        \toprule
        \textbf{Parameter} & \textbf{Value} \\
        \midrule
        River System & Mississippi River \\
        River Segments & 63 \\
        Total River Stations & 3,240 \\
        Average Station Spacing & 0.74 miles \\
        \bottomrule
    \end{tabular}
    \label{tab:dataset-overview}
\end{table}

\subsubsection*{ \textbf{Station Features}}
Each river station records key hydrodynamic and geometric attributes essential for river stage modeling:
\begin{itemize}
    \item \textit{Water Surface Elevation} – Height of the water column at a given station.
    \item \textit{Water Discharge} – Volume of water flowing per unit time.
    \item \textit{Geometric Information} – Cross-sectional profiles, bed elevation, and channel width.
\end{itemize}

\subsection{Baseline Model for Comparison}
The surrogate model is evaluated against the \textit{HEC-RAS numerical solver}, which serves as the \textbf{single-river ground truth} for model validation:

\begin{itemize}
    \item \textbf{HEC-RAS Numerical Solver:} The industry-standard method for solving the \textbf{Saint-Venant equations}, providing high-fidelity river stage predictions.
\end{itemize}

The evaluation focuses on:
\begin{itemize}
    \item Accuracy of water stage predictions compared to HEC-RAS.
    \item Computational efficiency gains from using a PINN-based surrogate model.
\end{itemize}

\subsection{Training Details}
The PINN model is implemented using TensorFlow and trained on an NVIDIA A100 GPU to leverage high-performance computing. The training follows a hybrid loss framework combining supervised learning with physics-informed constraints.

\begin{table}[htbp]
    \centering
    \caption{Training Configuration for Surrogate Model}
    \small
    \renewcommand{\arraystretch}{1.2} 
    \begin{tabular}{p{3.5cm} p{4.5cm}} 
        \toprule
        \textbf{Parameter} & \textbf{Description} \\
        \midrule
        Network Architecture & 6 hidden layers, 512 neurons/layer, ReLU activation \\
        Optimizer & Adam with learning rate \(10^{-3}\) and exponential decay \\
        Batch Size & 1024 samples per iteration \\
        Loss Weighting & Optimized via grid search \\
        Training Duration & 100,000 iterations \\
        \bottomrule
    \end{tabular}
    \label{tab:training-config}
\end{table}

\section{Benchmarking}
This section presents the benchmarking results comparing the execution time of a \textbf{HEC-RAS simulation} with a PINN-based surrogate model. The evaluation focuses on \textbf{1D unsteady flow analysis}, assessing computational efficiency and practicality. Given the significant computational demands of traditional hydrodynamic models like HEC-RAS, this benchmarking highlights the potential of surrogate models to deliver \textbf{faster simulations} while maintaining acceptable accuracy.

\subsection{Hardware Benchmarking}
Both the \textbf{HEC-RAS simulation} and the \textbf{per-river surrogate model} were executed on the same machine to ensure a direct comparison. The system specifications are:

\begin{table}[htbp]
    \centering
    \caption{Hardware Configuration for Benchmarking}
    \small
    \renewcommand{\arraystretch}{1.1} 
    \begin{tabularx}{\columnwidth}{X l} 
        \toprule
        \textbf{Component} & \textbf{Specifications} \\
        \midrule
        Operating System & Windows 11 Pro \\
        CPU & Intel Xeon E5-1620 v3 @ 3.5GHz \\ 
        & (4 cores, 8 threads) \\
        GPU & NVIDIA GeForce GTX 970 \\ 
        & CUDA 12.6, 4GB VRAM \\
        System Memory & 32GB RAM \\
        \bottomrule
    \end{tabularx}
    \label{tab:hardware-specs}
\end{table}

This consistent hardware configuration ensures that performance differences arise solely from variations in \textbf{software frameworks} and \textbf{algorithmic approaches}.

\subsection{Software Frameworks}
\subsubsection{Surrogate Model (Python-Based)}
The surrogate model is implemented in Python and leverages \textbf{GPU acceleration}. The implementation uses \textbf{Python 3.10.11}, with \textbf{PyTorch 2.1.1+cu121} for deep learning and \textbf{Torch Geometric 2.5.3} for graph-based computations.

\vspace{1em}
\subsubsection{HEC-RAS (CPU-Based)}
HEC-RAS serves as the baseline model for 1D unsteady flow analysis. The simulation was conducted using \textbf{HEC-RAS version 5.0.1}.

\subsection{Benchmarking Results}
The execution times for both methods are summarized below:

\begin{table}[htbp]
    \centering
    \caption{Total Execution Time Comparison (Seconds)}
    \small
    \begin{tabular}{l r} 
        \toprule
        \textbf{Simulation Task} & \textbf{Total Time (sec)} \\
        \midrule
        HEC-RAS Simulation (1D Unsteady Flow) & 8317 \\
        Surrogate Model (Per-River PINN) & 82.9 \\
        \bottomrule
    \end{tabular}
    \label{tab:total-execution-times}
\end{table}

Since each river segment model is run sequentially in a similar process to HEC-RAS, the execution time comparison remains consistent across both methods.

\begin{table}[htbp]
    \centering
    \caption{Phase Breakdown of HEC-RAS Execution Time (in Seconds)}
    \small
    \begin{tabular}{l r} 
        \toprule
        \textbf{Phase} & \textbf{Time (s)} \\
        \midrule
        Completing Geometry & <1 \\
        Preprocessing Geometry & 96 \\
        Unsteady Flow Computations & 8039 \\
        Writing to DSS & 7 \\
        Post-Processing & 172 \\
        \bottomrule
    \end{tabular}
    \label{tab:hec-ras-breakdown}
\end{table}

\subsubsection{Performance Comparison}
The \textbf{Per-River PINN} achieves a \textbf{100-fold performance improvement} compared to HEC-RAS. While the HEC-RAS simulation required \textbf{over two hours} (\textbf{8317 seconds}), the surrogate model completed the same analysis in \textbf{under 90 seconds} (\textbf{82.9 seconds}).

\subsubsection{Hardware Utilization}
\begin{itemize}
    \item \textbf{Surrogate Model:} The surrogate model benefits from \textbf{GPU acceleration}, enabling scalability with more powerful or additional GPUs.
    \item \textbf{HEC-RAS:} HEC-RAS is constrained by \textbf{CPU performance} and lacks GPU support, limiting scalability for larger or more complex simulations.
\end{itemize}

\subsubsection{Real-Time Applications}
These results highlight the viability of PINN-based surrogate models as a scalable and computationally efficient alternative to conventional numerical solvers for real-time river stage prediction.

\section{Results and Discussion}

\subsection{Evaluation Metrics}
To evaluate the performance of the single-river model, we consider the following key metrics:

\begin{itemize}
    \item \textbf{Mean Relative Absolute Error (MRAE)}: Measures prediction accuracy relative to observed river stage values, providing a scale-invariant assessment.
    \item \textbf{Physics Residual Loss}: Quantifies the model’s adherence to the governing physics equations.
    \item \textbf{Inference Time}: Assesses computational efficiency for potential real-time applications.
\end{itemize}

These metrics provide a more robust assessment of predictive accuracy, physical consistency, and computational feasibility than absolute error metrics.

\subsection{Single-River Model Evaluation}
The single-river models are evaluated independently on each river segment. The primary goal is to assess how well the PINN-based surrogate model can approximate HEC-RAS predictions without requiring retraining on other rivers.

\subsubsection{Stage Prediction Results}
Across all river stations, the model achieves a \textbf{low mean relative absolute error (MRAE)}, indicating strong predictive accuracy. Some rivers, such as the Tensas River, exhibit particularly low errors, while others, such as the Arkansas River, show slightly increased deviations.

\begin{figure}[htbp]
    \centering
    \includegraphics[width=0.7\linewidth]{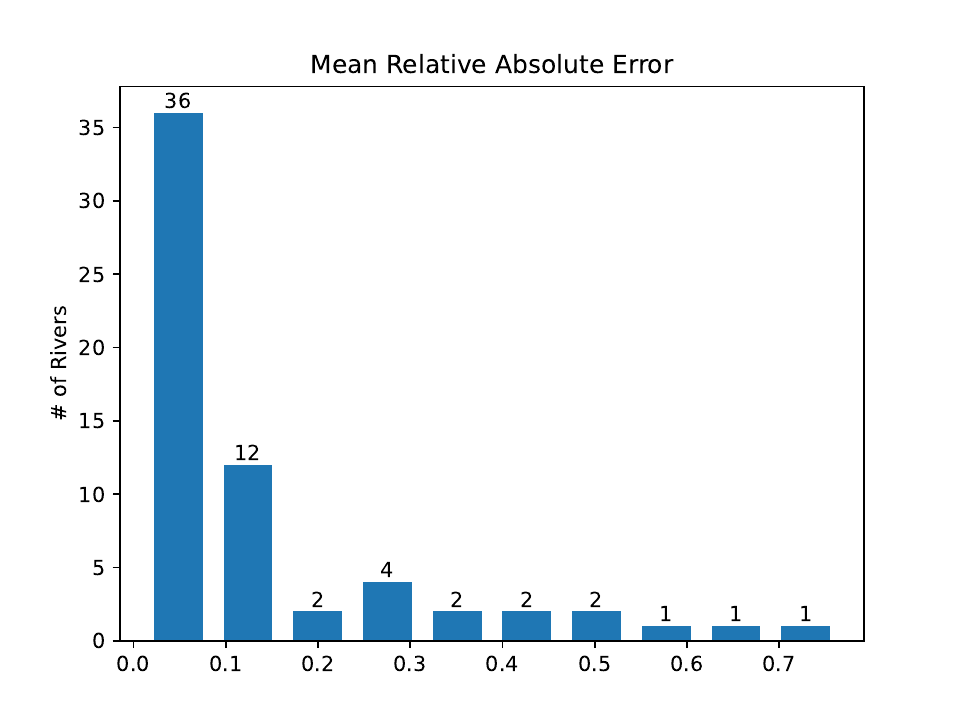}
    \caption{Histogram of relative error scores across river stations for the single-river model.}
    \label{fig:single-river-error}
\end{figure}

Figure~\ref{fig:single-river-error} depicts a histogram of the relative errors across different river stations. The majority of stations maintain low error rates, demonstrating the model’s ability to accurately capture river dynamics.

To further quantify model accuracy, we compute the \textbf{Mean Relative Absolute Error (MRAE)} as:

\begin{align}
    MRAE &= \frac{\frac{1}{N} \sum_{i=1}^{N} (|\hat{y_i} - y_i|)}{\frac{1}{N} \sum_{i=1}^{N} y_i}
    \label{eq:mean-relative-absolute-error}
\end{align}

\noindent where $\hat{y}$ is the predicted stage and $y$ is the ground truth.

Instead of absolute errors, this metric provides a relative measure of accuracy that accounts for variations in river stage values.

\subsection{Ablation Study}
To analyze the effectiveness of key model components, an ablation study was conducted by modifying:

\begin{itemize}
    \item \textbf{Random Fourier Features:} Improves the model’s ability to capture fine-scale variations in river stage.
    \item \textbf{Physics-Informed Regularization:} Ensures physically consistent water surface elevation predictions.
\end{itemize}

\begin{figure}[htbp]
    \centering
    \subfloat[Base Model (No Fourier Features, No Regularization)]{
        \includegraphics[width=0.32\textwidth]{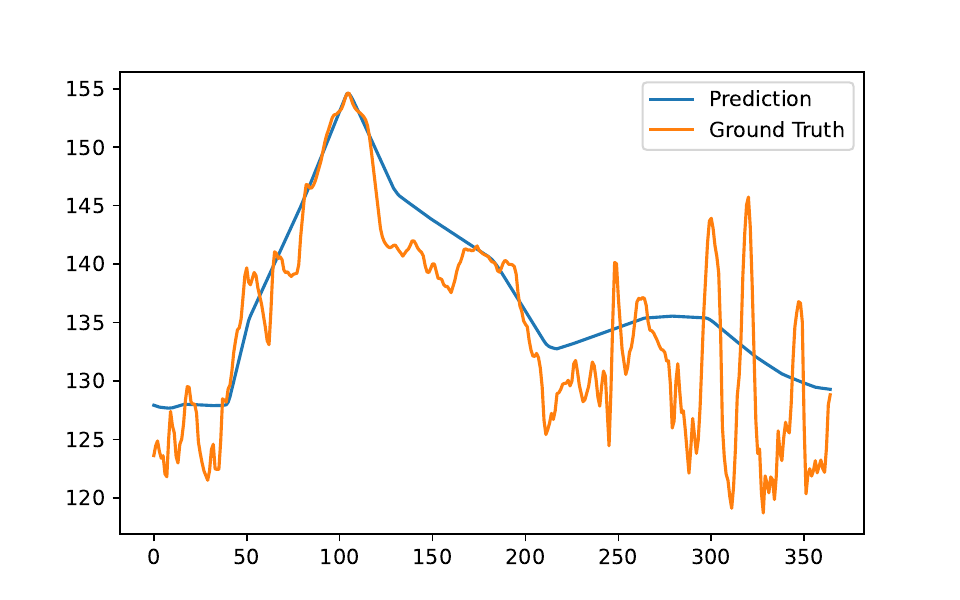}
        \label{fig:ablation-study-without-pinn-fourier}
    }
    \hfill
    \subfloat[With Random Fourier Features]{
        \includegraphics[width=0.32\textwidth]{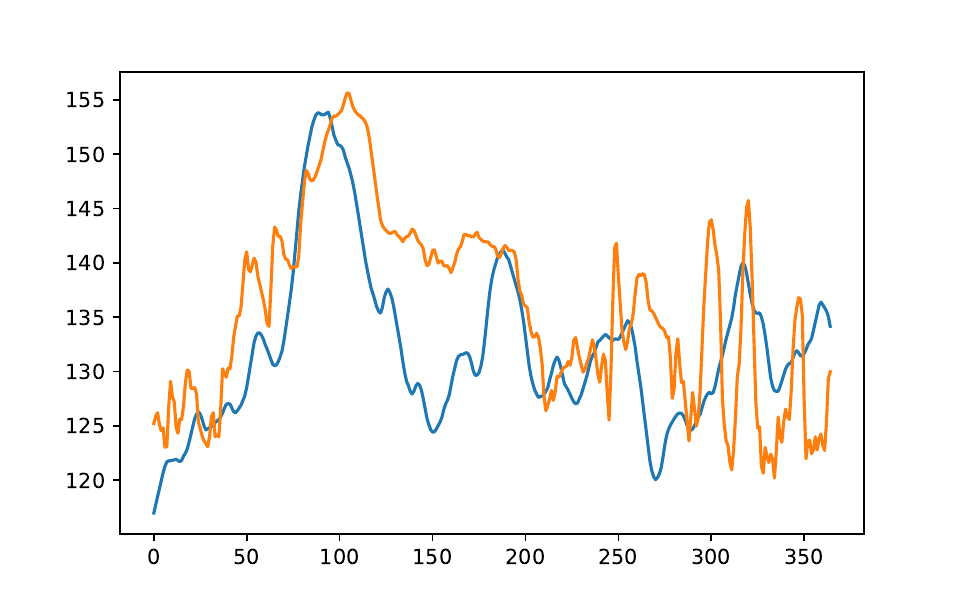}
        \label{fig:ablation-study-without-pinn}
    }
    \hfill
    \subfloat[With Physics-Informed Regularization]{
        \includegraphics[width=0.32\textwidth]{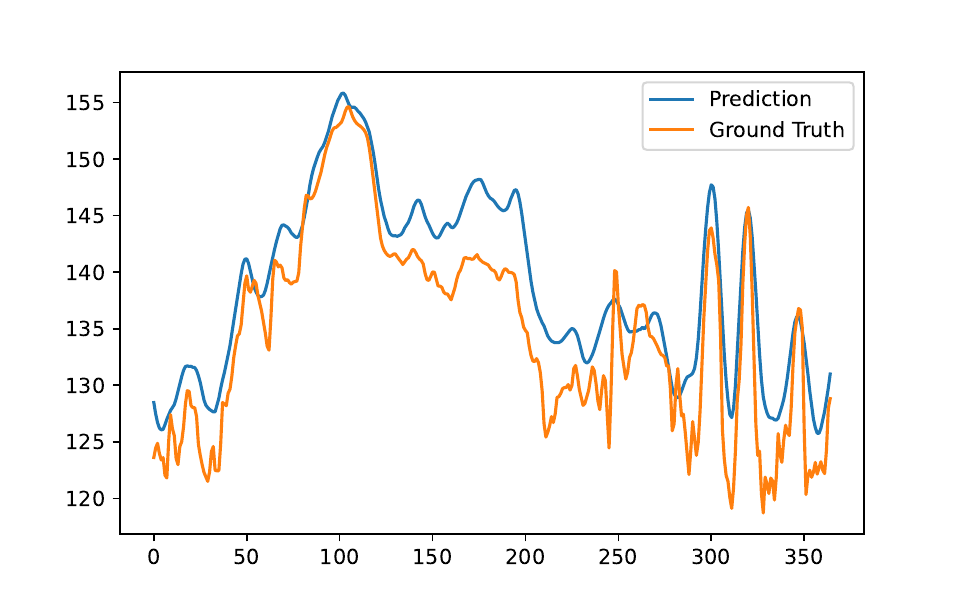}
        \label{fig:ablation-study-full}
    }
    
    \caption{Ablation study results demonstrating the impact of Fourier features and physics-informed regularization.}
    \label{fig:ablation-study}
\end{figure}

The results in Figure~\ref{fig:ablation-study} indicate:

\begin{itemize}
    \item Without Fourier features, the model struggles to capture high-frequency variations, resulting in a low-resolution approximation of the river stage.
    \item With Fourier features enabled, the model better captures small-scale fluctuations, although it may introduce noise if not regularized.
    \item Physics-informed regularization significantly improves global consistency, ensuring that the model adheres to known hydrodynamic principles.
\end{itemize}

\subsection{Findings and Discussion}
The evaluation of single-river models demonstrates three key findings:

\begin{itemize}
    \item \textbf{Single-river PINNs provide relatively accurate approximations of HEC-RAS stage predictions}, achieving a low mean relative error across most river stations.
    \item \textbf{Fourier feature encoding significantly enhances model resolution}, but requires physics-informed regularization to prevent overfitting.
    \item \textbf{The PINN-based model offers a computationally efficient alternative to HEC-RAS}, reducing inference time while maintaining physical consistency.
\end{itemize}

\section{Conclusion}

This study demonstrates that single-river PINNs can serve as computationally efficient surrogate models for river stage prediction, offering significant speed improvements over HEC-RAS while maintaining accuracy. To enhance applicability, future work should focus on generalizing the approach to multiple rivers and leveraging ensemble models for improved accuracy.

\subsection{Toward a Generalized Multi-River Model}

A key challenge remains in extending the current PINN framework to multiple river systems without requiring individual model retraining. Two promising strategies for generalization are:

\begin{itemize}
    \item \textbf{Geometry Encoding for Generalization}: Developing robust representations, such as cross-sectional encodings or graph-based models, to capture the diversity of river geometries \cite{Abdelguerfi2012}.
    \item \textbf{Ensemble Models for Robustness}: Combining multiple PINN models through weighted fusion, adaptive selection, or hybrid meta-models to improve predictive performance across different river segments.
\end{itemize}

\subsection{Future Research Directions}

Beyond generalization, additional research should focus on:

\begin{itemize}
    \item \textbf{Scaling PINNs for 2D/3D Hydrodynamics}: Extending models to simulate multi-dimensional flow for complex river and coastal systems.
    \item \textbf{Real-World Validation}: Comparing PINN predictions with observed river stage data to refine accuracy.
    \item \textbf{Adaptive Loss Weighting}: Optimizing the balance between physics-informed constraints and data-driven loss functions for improved model stability.
\end{itemize}

By advancing these strategies, PINNs can evolve into scalable, physics-consistent, and computationally efficient tools for river stage forecasting, reducing reliance on traditional numerical solvers while enabling real-time hydrodynamic predictions.


\end{document}